\documentclass[journal,comsoc]{IEEEtran}
\usepackage{mmstyles}
\usepackage{graphicx}
\usepackage{comment}
\usepackage{hyperref}
\usepackage{microtype}
\usepackage{subcaption}
\usepackage{multirow}
\usepackage{color}
\usepackage{algpseudocode}
\usepackage{algorithm}
\usepackage{amsmath}

\usepackage[T1]{fontenc}

\newcommand{\methodname}{PM-Net}

\newcommand{\newpara}[1]{\noindent\textbf{#1}}

\ifCLASSINFOpdf
\else
\fi
\usepackage{amsmath}
\usepackage[cmintegrals]{newtxmath}
\begin{document}
%
\title{Chasing the Tail in Monocular 3D Human Reconstruction with Prototype Memory}
%
%
%

\author{Yu~Rong, Ziwei~Liu and Chen~Change~Loy}
\twocolumn[{
	\renewcommand\twocolumn[1][]{#1}
	\maketitle
	\begin{center}
		\centering
		\includegraphics[width=\linewidth]{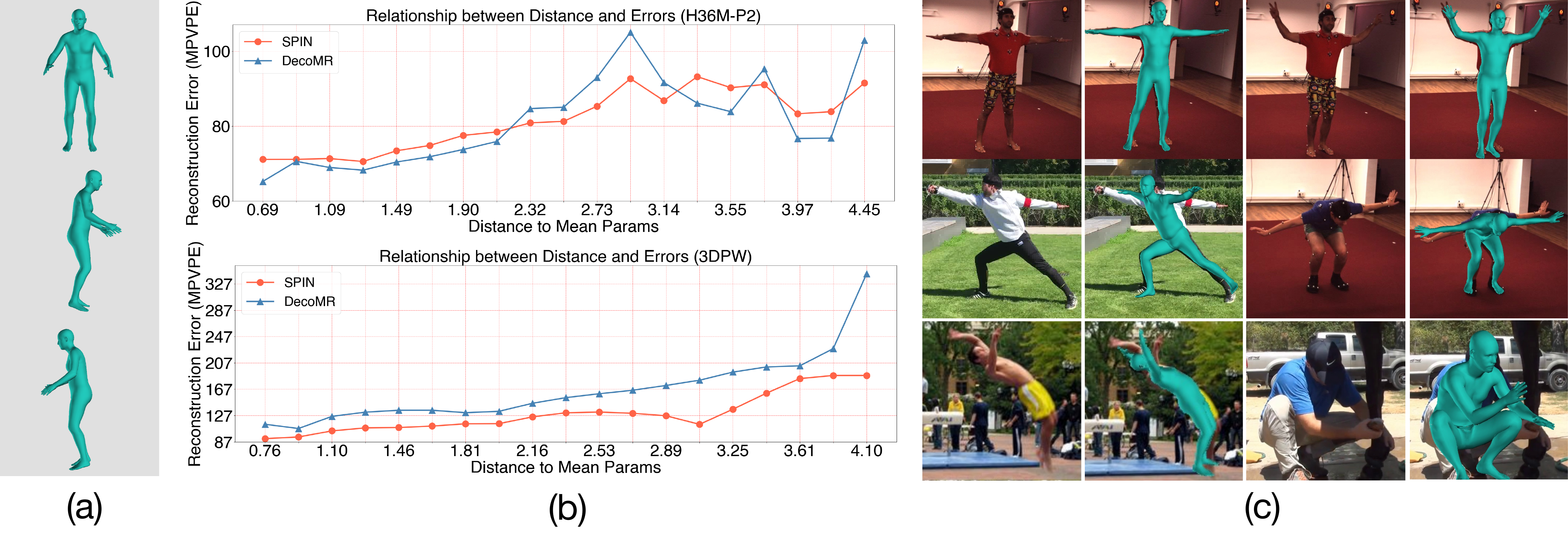}
		\captionof{figure}{\textbf{Overview}. In (a), we show one typical singular human prototype used by previous works~\cite{kolotouros2019spin}. In (b), we plot the relationship between the samples' distances to the mean parameters and models' reconstruction error (mean per-vertex position error) on the two widely used evaluation sets, ~\ie the evaluation sets of Human3.6M~\cite{ionescu2014human3} and 3DPW~\cite{vonMarcard2018}. We select two state-of-the-art 3D human reconstruction models, namely SPIN~\cite{kolotouros2019spin} and DecoMR~\cite{zeng20203d}. In (c), we show several typical examples with gradually increasing distances to the human prototype depicted in (a). As the figure shows, the model (SPIN~\cite{kolotouros2019spin}) generates less precise results on samples with larger distances.}
		\label{fig:teaser}
	\end{center}
}]

\begin{abstract}
Deep neural networks have achieved great progress in single-image 3D human reconstruction. However, existing methods still fall short in predicting rare poses.
The reason is that most of the current models perform regression based on a single human prototype, which is similar to common poses while far from the rare poses.
In this work, we 1) identify and analyze this learning obstacle and 2) propose a prototype memory-augmented network, \methodname, that effectively improves performances of predicting rare poses.
The core of our framework is a memory module that learns and stores a set of 3D human prototypes capturing local distributions for either common poses or rare poses.
With this formulation, the regression starts from a better initialization, which is relatively easier to converge.
Extensive experiments on several widely employed datasets demonstrate the proposed framework's effectiveness compared to other state-of-the-art methods.
Notably, our approach significantly improves the models' performances on rare poses while generating comparable results on other samples. 

\end{abstract}


%
\IEEEpeerreviewmaketitle

\section{Introduction}
\label{sec:intro}

Recovering 3D human model from a single monocular image can facilitate numerous applications in augmented reality and creative entertainment.
Most existing methods~\cite{kolotouros2019spin,zeng20203d,kanazawa2018end,guler2019holopose,kolotouros2019cmr}
employ a parametric 3D human shape model known as Skinned Multi-Person Linear Model (SMPL)~\cite{loper2015smpl} to represent 3D human and use deep convolutional neural networks (DCNN) to estimate its parameters. 
They also employ 3D human prototype, such as mesh template~\cite{zeng20203d,kolotouros2019cmr} or mean parameters~\cite{kolotouros2019spin,kanazawa2018end}, as the foundation to predict 3D human models.
Using these methods, images with common poses are satisfactorily predicted. However, performances of these models decline drastically when applied to images with rare poses or uncommon views, as shown in the third row of Fig.~\ref{fig:teaser} (c).

The performance degradation in predicting rare poses is primarily caused by the usage of a single 3D human prototype for all the samples.
%
%
The single human prototype calculated from these datasets, as visualized in Fig.~\ref{fig:teaser} (a), is more similar to common poses with usual viewpoints, such as standing.
Models trained with this single prototype inherently bias towards the common poses while performing less well on rare poses.
To illustrate this phenomenon, we show the visualization of the single human prototype used by SPIN~\cite{kolotouros2019spin} in Fig~\ref{fig:teaser} (a) and calculate the Euclidean distance of each sample to the human prototype. 
We use the distances between the corresponding SMPL vertices.
As depicted in Fig.~\ref{fig:teaser} (b), when a single prototype is adopted, two state-of-the-art models, namely SPIN~\cite{kolotouros2019spin} and DecoMR~\cite{zeng20203d} perform well on samples that are close to the human prototype. However, their reconstruction errors rise rapidly on samples that have a larger distance to the mean parameters.

Take SPIN~\cite{kolotouros2019spin} and Human3.6M~\cite{ionescu2014human3} for instance, the reconstruction errors of samples with distance to mean parameters of around 4.3 is 20\% larger than those with distance to mean parameters of around 2.0.
The performance drop is more severe on challenging in-the-wild datasets such as 3DPW~\cite{vonMarcard2018}.
We further show several typical examples in Fig.~\ref{fig:teaser} (c). It is observed that current state-of-the-art models' prediction becomes imprecise when the samples' poses gradually differ from the human prototype.
For better clarification, in the following of the paper, we call the samples that are close to the human prototype head classes and the samples far from the human prototype tail classes. Typically, head classes are composed of common poses while rare poses majorly lie in tail classes.

To ameliorate the models' performance collapse on tail classes, we propose a prototypical memory network, \methodname. 
The core of {\methodname} is a memory module that stores multiple 3D human prototypes, each of which covers a non-overlapping subset of the data. 
Instead of using one single prototype for all data samples, we assign each data to the closest prototype in the memory.
In this way, the distance of data to its assigned prototype is significantly reduced. 
The regression process starts from a much better initialization.
Furthermore, the regression process is transformed to be performed on compact local distributions instead of on
the global yet sparse distribution.
%
%
During inference, a classifier is employed to assign input data to the corresponding prototype, 
which will serve as the basis for SMPL parameter regression.

Previous works such as LCR-Net~\cite{rogez2019lcr} also adopts multiple human prototypes by applying standard K-Means clustering on 3D poses data. 
Nevertheless, it is not feasible to directly apply the same method in 3D human reconstruction. 
Firstly, although the distances between 3D poses can be easily measured by Euclidean distance between coordinates, it is not reasonable to use the same metric on SMPL parameters. Distances between vertices are more reasonable. 
Furthermore, the weights of each vertex should be elaborately selected in calculating the distance. 
For example, the number of vertices that belong to the head is nearly the same as the number of vertices of limbs, 
while the latter has much more influence on the overall poses and thus should be placed larger weights during the distance calculation.
Another issue is how to obtain the centers of each cluster after clustering. 
A direct average of vertices is obviously infeasible, since the resulting vertices are highly likely to not lie in the valid SMPL topology. 
To solve this issue, we choose to conduct on the samples' SMPL parameters and apply separate averaging strategies on pose and shape parameters to suit their properties better.

In summary, we make the following contributions:  
We identify the relationship between monocular 3D human reconstruction models' performances and samples' distance to the applied singular human prototypes. 
To improve the models' performance on tail classes, we design a prototype memory module to fully leverage the information contained across all samples of common and rare poses. 
Thanks to the unique formulation of local prototypes, our method improves the models' performances on both tail classes and the overall samples.
Our model achieves state-of-the-art performances on Human3.6M~\cite{ionescu2014human3}, MPI-INF-3DHP~\cite{mehta2017monocular}, 3DPW~\cite{vonMarcard2018} and UP-3D~\cite{lassner2017unite}. 
In particular, the model's average reconstruction errors on tail classes are reduced by 12 mm on challenging in-the-wild dataset, ~\ie UP-3D~\cite{lassner2017unite}.
The proposed method is easy to implement and can be adapted to various frameworks or other visual tasks that perform regression from mean parameters.

\section{Related Work}
\label{sec:related_work}

\newpara{Single-Image 3D Human Reconstruction.}
Recent methods for single-image 3D human reconstruction share a similar pipeline. 
In particular, a parametric model SMPL~\cite{ionescu2014human3} is used to represent 3D human, while DNN or optimization is employed to obtain the parameters or vertex coordinates of the SMPL model.
Most of the learning-based methods~\cite{kanazawa2018end,pavlakos2018learning,xu2019denserac,rong2019delving,moon2020i2l,georgakis2020hierarchical,zanfir2020weakly} use CNN-based models to predict the parameters of SMPL directly.
Kanazawa~\etal~\cite{kanazawa2018end} use adversarial losses~\cite{goodfellow2014generative} to judge the predicted 3D human poses.
%
%
Pavlakos~\etal~\cite{pavlakos2018learning} propose to predict the 2D poses heatmaps and silhouette as the intermediate representation to facilitate the model prediction.
G\"uler~\etal~\cite{guler2019holopose}, Xu~\etal~\cite{xu2019denserac} and Rong~\etal~\cite{rong2019delving} incorporate DensePose~\cite{densepose} into the framework.
Moon~\etal~\cite{moon2020i2l} design a framework to predict 3D poses first and regress the corresponding SMPL parameters based on the predicted 3D poses.
Georgakis~\etal~\cite{georgakis2020hierarchical} adopt a hierarchical strategy to predict the SMPL poses part by part, instead of regressing all together.
Other learning-based methods adopt different frameworks. 
Kolotouros~\etal~\cite{kolotouros2019cmr} and Choi~\etal~\cite{choi2020pose2mesh} use graph convolution network~\cite{kipf2017semi} to predict SMPL vertices directly.
Zeng~\etal~\cite{zeng20203d} regress location maps in the UV map space instead of the original SMPL parameter space.

Different from the aforementioned learning-based methods, optimization-based methods directly optimize the SMPL parameters instead of using DNNs. 
Bogo~\etal~\cite{bogo2016keep} obtains SMPL parameters by minimizing the distances between ground-truth 2D keypoints and projected 2D keypoints from predicted SMPL models.
%
%
Kolotouros~\etal~\cite{kolotouros2019spin} designs a framework to unify the learning-based and optimization-based methods. 
They first deploy a CNN model to predict the SMPL parameters of the given images. 
The predicted SMPL parameters then serve as the initialization of optimization fitting~\cite{bogo2016keep}.
After that, the optimized SMPL parameters are used to supervise the CNN model. Those steps are performed in iteration until convergence.
Our model shares a similar framework as SPIN~\cite{kolotouros2019spin} while we use the fitted SMPL parameters to supervise the model instead of performing ``in-the-loop`` optimization in the training process.
Another major difference is that we apply prototypical memory and assign each sample the closest prototype learned from data.

\begin{figure*}[t]
	\begin{center}
		\includegraphics[width=1.0\linewidth]{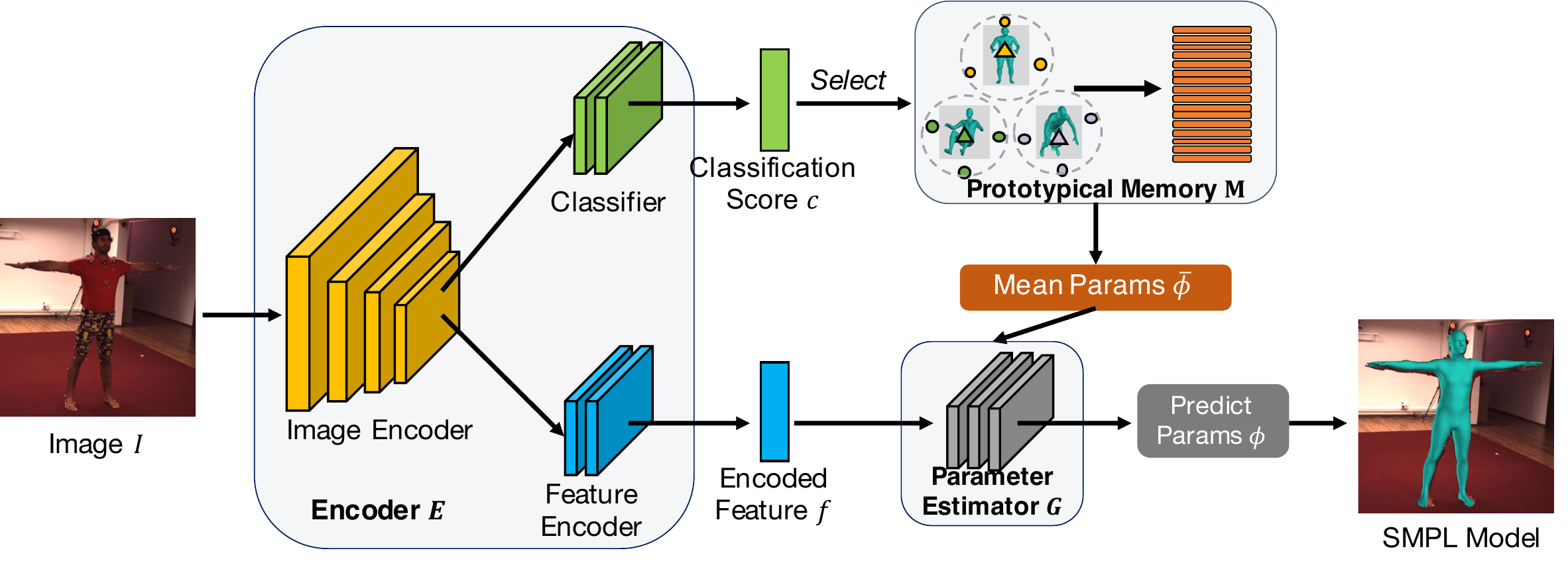}
	\end{center}
	\vskip -0.3cm
	\caption{\textbf{Overall framework of \methodname}. The framework is composed of an encoder $E$, a parameter estimator $G$ and a prototypical memory $M$. The Encoder $E$ is composed of an image encoder, a feature encoder and a classifier. The most important component that distinguishes our work from other methods is the Prototypical Memory $M$ that stores multiple sets of mean parameters. Using the classification score $c$ generated by the encoder, it selects the matching mean parameters $\bar{\phi}$. Parameter Estimator $G$ uses the matching mean parameters as the initialization and regresses the SMPL parameters $\phi$ after several iterations. In the end, SMPL model generates the corresponding vertices and joints with the predicted parameters.}
	\label{fig:pipeline}
	\vspace{0.1cm}
\end{figure*}

\newpara{Using Prototypes in 3D Human Reconstruction.}
Most of the learning-based methods employ a 3D human prototype as the foundation to predict the required values.
For example, several HMR~\cite{kanazawa2018end}-based methods~\cite{kolotouros2019spin,xu2019denserac,rong2019delving} use mean parameters calculated from datasets to represent the 3D human prototype.
G{\"u}ler~\etal~\cite{densepose} use a set of Euler angles to form a convex hull as the basis for SMPL pose prediction.
Kolotouros~\etal~\cite{kolotouros2019cmr} use mesh template as the input of a graph convolutional network~\cite{kipf2017semi} to directly regress 3D vertices of the SMPL model.
DecoMR~\cite{zeng20203d} also adopts the reference mesh as the base for predicting location maps.
Although these works achieve promising results on common poses, their performances drop drastically when applied to images with rare poses. 
It is because such methods only adopt one single 3D human prototype that is close to common poses while far from rare poses. Therefore, these methods are biased towards predicting common poses.
To overcome this drawback, we propose to use multiple 3D human prototypes. Each data sample is assigned to the learned prototype that it is closest to.
To our knowledge, LCR-Net~\cite{rogez2019lcr} is the only work that applies multiple prototypes for 3D human estimation area. The prototypes used by LCR-Net is obtained from obtaining naive K-Means to 3D pose data.
Nevertheless, obtaining effective prototypes for 3D human reconstruction is not that trivial. Directly applying naive K-Means on SMPL parameters,~\ie, pose parameters and shape parameters, will lead to suboptimal performance.
To circumvent this hurdle, we elaborately design a specified clustering algorithm for SMPL model, which is more effective in 3D human reconstruction.

\section{Methodology}
\label{sec:approach}

\begin{table}[t] \centering  
	\caption{List of mathematical symbols.}
	\begin{tabular}{c|c}
		\hline
		Meaning  & Math Symbol \\
		\hline
		SMPL Pose Parameters & $\theta$ \\
		SMPL Shape Parameters & $\beta$ \\
		SMPL Parameters & $\phi = (\theta, \beta)$ \\
		SMPL Vertices & $V$ \\
		SMPL Joint Regressor & $J$ \\
		3D Joints & $J^{3D}$ \\
		2D Joints & $J^{2D}$	 \\
		Keypoints Visibility & $\mu$ \\
		\hline
		Cluster Center & $\pi$ \\
		Weights for each Vertex & $W$ \\
		\hline
		Input Image & $I$ \\
		Encoder & $E$ \\
		Parameter Estimator & $G$ \\
		Prototypical Memory & $M$ \\
		Encoded Feature & $f$ \\
		Classification Score & $c$ \\
		Number of Clusters & $K$ \\
		\hline
	\end{tabular}	
	\label{tab:math_symbol}
\end{table}

To improve the models' performances on tail classes, we propose a general framework called Prototypical Memory Network (\methodname), as depicted in Fig.~\ref{fig:pipeline}. 
The framework is composed of three parts: an encoder, a parameter estimator and a prototypical memory.
The core component is a prototypical memory that stores multiple 3D human prototypes.
Follow the practices of previous works~\cite{kolotouros2019spin,kanazawa2018end}, we use mean SMPL parameters as the 3D human prototype.
During inference,
the encoder takes images as input and outputs feature vectors and prototype classification scores. The prototype classification scores are used to select mean parameters from the prototypical memory. 
After that, the feature vectors and selected mean parameters are fed into the parameter estimator. 
Lastly, the parameter estimator uses the selected mean parameters as the initialization and estimates the predicted parameters in an iterative fitting manner~\cite{kolotouros2019spin,kanazawa2018end}.

In the following sections, we will first introduce the 3D human model used in this paper and the inference process of ~\methodname. 
Next, we present the design of prototypical memory. Finally, we discuss the training strategy used in this work.
The mathematical symbols used in this paper are listed in Table~\ref{tab:math_symbol}.

\subsection {3D Human Model}
\label{subsec:smpl}
We use SMPL~\cite{loper2015smpl}, a triangle-mesh-based model parameterized by the shape and pose parameters to represent a 3D human model.
The shape parameters $\boldsymbol{\beta} \in \mathbb{R}^{10}$ influence the overall body shape.
The pose parameters $\boldsymbol{\theta}$ model both the relative 3D rotations for a predefined kinematics skeleton with $23$ joints and the global rotation for the whole body.
The original SMPL pose parameters are in angle-axis representation. Follow the practice of SPIN~\cite{kolotouros2019spin}, we use the continuous rotation representations proposed by Zhou~\etal~\cite{zhou2019continuity} to represent 3D rotations.
Therefore, the dimension of pose parameters becomes $\boldsymbol{\theta} \in \mathbb{R}^{6 \times 24}$.
Given the shape and pose parameters, SMPL model calculates the coordinates of vertices $\boldsymbol{V} \in \mathbb{R}^{3 \times 6890}$. 
Given SMPL vertices $\boldsymbol{V}$, 3D joints $\boldsymbol{J^{3D}} \in \mathbb{R}^{3 \times 24}$ are obtained by using predefined joint regressor $\boldsymbol{J} \in \mathbb{R}^{24 \times 6890}$ by $J^{3D} = J \cdot V^{T}$.
%
%

\subsection{Framework}

\begin{figure}[t]
	\begin{center}
		\includegraphics[width=0.9\linewidth]{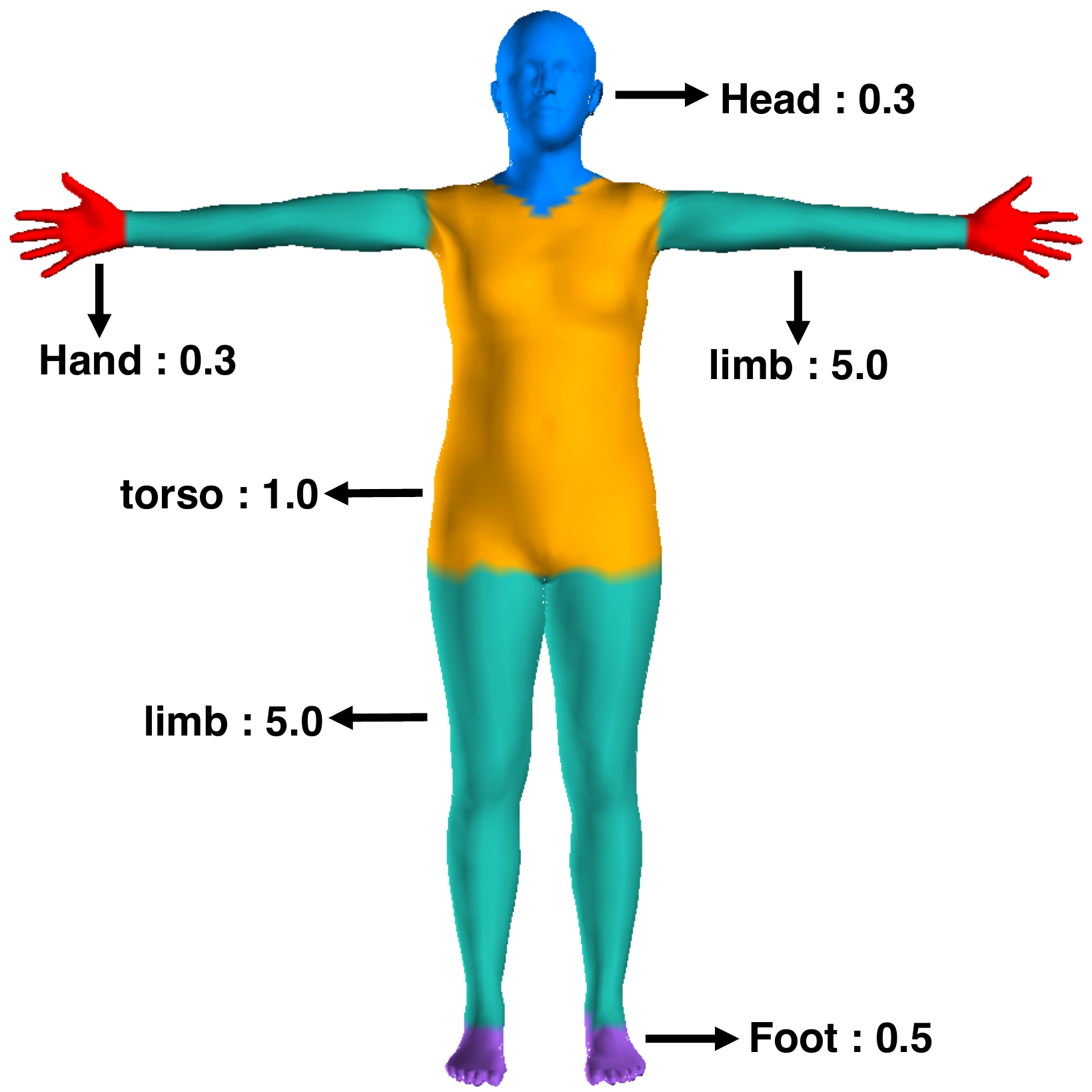}
	\end{center}
	\caption{\textbf{Visualization of vertices belonging to different body parts.} In performing part-aware weighting, we divide whole body into different body parts and assign different weights for each part. The body parts include head, hand, foot, limbs (arms and legs) and torso.}
	\label{fig:body_part}
\end{figure}

The overall framework is shown in Fig.~\ref{fig:pipeline}. A single image $I$ serves as the only input.
The encoder $E$, composed of an image encoder, a feature encoder and a classifier, takes in input $I$ and outputs encoded feature $f$ and classification score $c$. 
The score $c$ is then used to selected matching mean parameters $\bar{\phi} = (\bar{\theta}, \bar{\beta})$ from the memory $M$.
Given the encoded feature $f$ and selected mean parameters $\bar{\phi}$, the parameter estimator $G$ regresses the SMPL parameters $\phi$. 
In the last step, 3D meshes are generated by the SMPL model using the estimated parameters.

%
Contrary to the previous practices~\cite{kolotouros2019spin,kanazawa2018end} that use the same mean parameters for all samples, our model assigns different sets of mean parameters to different samples.
The framework maintains a prototypical memory that is composed of multiple sets of mean SMPL parameters.
Given each sample, the matching set of parameters is selected. 
The number of mean parameter sets is fixed as $K$. These $K$ set of mean parameters further form the mean parameters memory $M \in \mathbb{R}^{K \times D}$, where $D$ is the dimension of the parameters. 
In data preprocessing, each data sample is assigned a one-hot class label $\hat{c} \in \mathbb{R}^{1 \times K}$, indicating to which set of mean parameters the sample is closest to.
During inference, the classifier in the encoder $E$ produces a classification score $c \in \mathbb{R}^{1 \times K}$ for each sample.
With the classification score $c$, the mean parameters $\bar{\phi}$ are selected from the prototype memory $M$ as $\bar{\phi} = cM$.
The obtained $\bar{\phi}$ serves as the initialization for iteratively fitting the parameters as $\phi_{t+1} = \phi_{t} + \Delta\phi_{t}$, where $\phi_{0} = \bar{\phi} $.
Follow the practice of previous works~\cite{kolotouros2019spin,kanazawa2018end}, fitting process is iterated for $3$ times.

\subsection {Building Prototypical Memory}
\label{sec:prototypical_memory}


\begin{algorithm}
	\caption{Part-Aware 3D Human K-Means}\label{p3dh_kmeans}
	\begin{algorithmic}[1] 
		\Require Threshold for average sample-to-center distance $\hat{\gamma}$; 
		\Require Threshold for total number of iterations $\hat{\lambda}$.
		\Require N samples: $\Phi = \{\phi_1, ... \phi_N \}$; $\phi_i \in \mathbb{R}^{154}$
		\Require Initial clusters centers: $\Pi = \{ \pi_1, ..., \pi_K\}$; $\pi_j \in \mathbb{R}^{154},$
		\Require Weights for each vertex $W \in \mathbb{R}^{3 \times 6890}$. 

		\Procedure{clustering}{$\hat{\gamma}$, $\hat{\lambda}$, $\Phi$, $\Pi$, $W$}
		
		\State Initialize sets of assigned samples $\Phi_{i} = \{\},\; i=1,\ldots,K$
		\State Initialize set of sample-to-center distances. $\Gamma = \{\}$.
		\State Initialize average sample-to-center-distance $\bar{\gamma} = \infty$; 
		\State Initialize the number of iterations $\lambda = 0$.
		
		\While {$\bar{\gamma} < \hat{\gamma}$ or $\lambda < \hat{\lambda}$}
		\State 
		\For {$i\gets 1,N$} \Comment{Assign samples.}
		\For {$j\gets 1,K$}
		\State $V_i = smpl(\phi_i)$; $V_j = smpl(\pi_j)$;
		\State $\gamma_{ij} = \lVert (V_i - V_j) \circ W \rVert^{2}_{2}$
		\EndFor
		\State $a_i = \mathop{\argmin}_{j} (\gamma_{ij}) $.
		\State $\gamma_i = \mathop{\argmin}_{\gamma_{ij}} (\gamma_{ij}) $.
		\State Add sample $\phi_i$ to $a_i$-th cluster.
		\State Add distance $\gamma_i$ to $\Gamma$.
		\EndFor
		
		\State 
		\For {$j\gets 1,K$} \Comment{Update cluster centers.}
		\State $\Theta_j = \{\}$, $B_j = \{\}$
		\For {$\phi_{jk}$ in $\Phi_{j}$}
		\State $(\theta_{jk}, \beta_{jk}) = \phi_{jk} $.
		\State Add $\theta_{jk}$ to $\Theta_j$.
		\State Add $\beta_{jk}$ to $B_j$.
		\EndFor
		\State $\theta_j = average\_\theta(\Theta_j) $.
		\State $\beta_j = average\_\beta(B_j) $.
		\State $\pi_j = (\theta_j, \beta_j)$.
		\EndFor
		
		\State 
		\State $\bar{\gamma} = average(\Gamma)$. \Comment{Update average distances.}
		\State $\lambda = \lambda + 1$.
		
		\EndWhile
		
		\EndProcedure

	\end{algorithmic}
\end{algorithm}

\begin{figure*}[t]
	\begin{center}
		\includegraphics[width=1\linewidth]{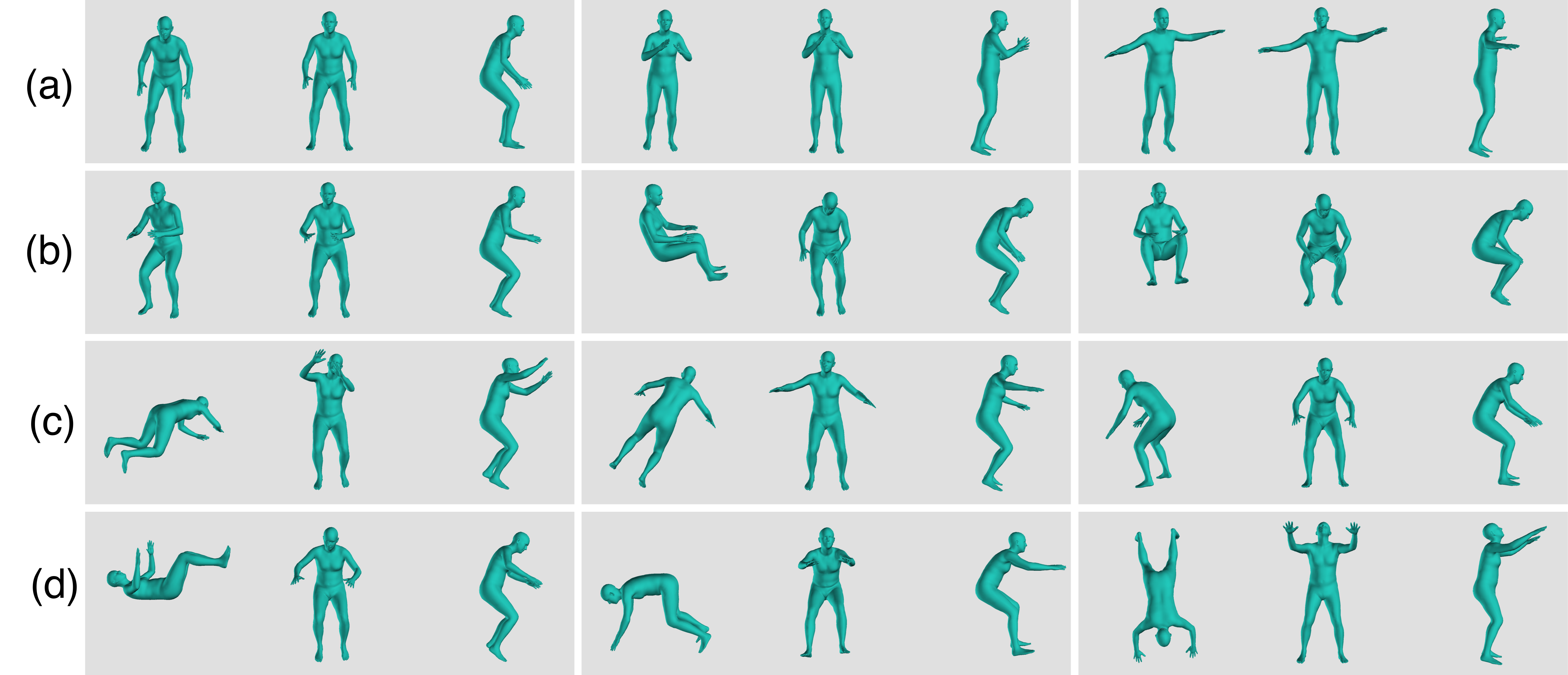}
	\end{center}
	\caption{\textbf{Visualization of 3D human prototypes in the prototypical memory.} We visualize several representative sets of mean parameters stored in the prototypical memory. We render each selected mean parameters in three views: the original view, frontal view and side view. From the figure, it can be found that the prototypical memory covers both commons poses like standing ((a)) and rare poses such as squatting ((b)). It also contains both common view types (frontal view in (a) and back view in (c)) and rare view types, such as head upside down in (d).}
	\label{fig:center_vis}
\end{figure*}

To construct the prototypical memory, we first apply clustering on training data. The obtained cluster centers are used to compose the memory. 
Instead of applying naive K-Means clustering algorithm on SMPL parameters $\phi = \left(\theta,\beta \right)$,
we elaborately design a clustering algorithm that exploits the characteristics of 3D human reconstruction. 
The algorithm is based on K-Means with following modifications:

\noindent
1) \textit{Distance calculating} -- Instead of calculating the distance between parameters $\phi$, we use the SMPL model to obtain the vertices $V$ of the corresponding parameters $\phi$. Then we calculate the Euclidean distance based on the vertex coordinates.
We argue that this distance format is more suitable since it directly and effectively reflects the pose and shape variations across different samples.
Furthermore, in calculating the distance based on vertices, we find it is suboptimal to assign the same weights to all the vertices. 
For example, the number of vertices belonging to the head is similar to the number of vertices of limbs, while the limbs have more influence on the overall pose estimation. 
Therefore, we assign larger weights to body parts that have more influences on pose estimation while smaller weights to the other parts. 
We call this strategy as \textit{part-aware weighting}.
In our experiment, the weights for vertices belonging to different body parts are empirically set as $5.0$ for limbs (arms and legs), $0.3$ for head and hand, $0.5$ for foot, $1.0$ for torso.
Visualization of vertices belonging to different body parts is shown in Fig.~\ref{fig:body_part}.

\noindent
2) \textit{Cluster center updating} --
To update centers of each cluster after each iteration, a straightforward idea is to directly average pose and shape parameters of the assigned samples. 
Averaging shape parameters is valid since the process of blending shapes is linear.
Nevertheless, directly averaging pose parameters, which are in the form of 3D rotations, is not reasonable, as pointed out by previous works~\cite{markley2007averaging}.
To obtain the valid averaged rotations, we first transfer the 3D rotations to the format of quaternions. 
Then we apply the quaternion averaging algorithm proposed by Markley~\etal~\cite{markley2007averaging}.

To be more specific, suppose we have $n$ quaternions $q_i$ to be averaged, the averaged quaternions $\bar{q}$ can be obtained by solve the following equations:
\begin{equation}
\label{equ:ave_quat}
\begin{aligned}
  &	M = \sum_{n}^{i=1} q_i q_i^T \\
  & \bar{q} = \mathop{\argmax}_{q \in \mathbb{S}^3} q^T M q \\
\end {aligned}
\end{equation}

\noindent
$\mathbb{S}^3$ is the $3$ dimensional unit sphere. The solution of the maximization problem is the eigenvector of $M$ which corresponding to the maximum eigenvalue.
We refer the readers to the original paper~\cite{markley2007averaging} for more details.

Equipped with the aforementioned characteristics, we design a new clustering algorithm for 3D human reconstruction. We call it Part-Aware 3D Human K-Means (P3DH K-Means). 
The whole process of P3DH K-Means is composed of initialization, sample assignment and center updating. The number of clusters is fixed to be $K$. Suppose we use $N$ samples in performing the clustering. 
Each sample is represented as the concatenation of its pose parameters $\theta_i \in \mathbb{R}^{6 \times 24}$ and shape parameters $\beta_i \in \mathbb{R}^{10}$.
The centers of each cluster $\pi_j$ are represented in the same way. 
In the initialization step, we randomly select $K$ samples from the training data and use them as the initial cluster centers. 
In cluster assignment step, for each sample, the distances of its SMPL vertices between the SMPL vertices of each cluster center are calculated. The sample is then assigned to the closest cluster based on the distances.
In center updating step, shape parameters are calculated by directly averaging the shape parameters of the assigned samples.
To calculate the pose parameters, the pose parameters of the assigned samples are first converted to quaternions format. 
Then the quaternion averaging algorithm proposed in ~\cite{markley2007averaging} is used to obtain the averaged quaternions.
After that, the averaged quaternions are converted back to the continuous rotation representation~\cite{zhou2019continuity}.
The process of sample assignment and center updating iterates until the algorithm converges.
The whole process of P3DH K-Means is presented in Algorithm~\ref{p3dh_kmeans}. 
We further show several selected cluster centers in Figure~\ref{fig:center_vis}. 
It is observed that the obtained cluster centers cover both common and rare poses with different view types.

\subsection{Training Scheme}
We employ 3D and 2D losses to train our model. 
The 3D loss is composed of SMPL pose parameter loss $L_{\theta}$, SMPL shape parameter loss $L_{\beta}$ and 3D joint loss $L_{J3D}$. 
These losses are defined as follows:

\begin{equation}
\label{eq:3d_loss}
\begin{aligned}
& L_{\theta} = \lVert \theta - \hat{\theta} \rVert^{2}_{2}, \\
& L_{\beta} = \lVert \beta - \hat{\beta} \rVert^{2}_{2}, \\
& L_{J3D} =  \lVert J^{3D} - \hat{J}^{3D} \rVert^{2}_{2}, \\  
\end{aligned}
\end{equation} 
where $\hat{\theta}$, $\hat{\beta}$ and $\hat{J^{3D}}$ are ground truth SMPL pose parameters, shape parameters and 3D Joints.

To estimate 2D keypoints, we adopt a weak perspective camera model that is composed of scale $s$ and camera translation $[t_x, t_y]$. The camera parameters $[s, t_x, t_y,]$ are also predicted by the parameter estimator $G$. 
Given the camera parameters, 2D keypoints $J^{2D}$ are projected from $J^{3D}$. Then predicted $J^{2D}$ is used to calculate the 2D keypoint loss $L_{J2D}$ with the ground-truth 2D keypoints $\hat {J^{2D}}$.
The whole process is formulated as below,
\begin{equation}
\label{eq:2d_loss}
\begin{aligned}
& J^{2D}_{x} = s \times J^{3D}_{x} + t_x, \\
& J^{2D}_{y} = s \times J^{3D}_{y} + t_y, \\
& L_{J2D} = \lVert \mu \cdot (J^{2D} - \hat{J^{2D}}) \rVert^{2}_{2}, \\
\end{aligned}
\end{equation}
where $\mu$ is the visibility indicator.

Since our model requires assigning samples into different prototypes, we apply a cross-entropy classification loss $L_C$, defined as:
\begin{equation}
\label{eq:classify_loss}
\begin{aligned}
& L_{C} = -\hat{c} \cdot \log(c). \\
\end{aligned}
\end{equation}

\noindent
\textbf{Overall Loss Function.}
The overall loss $L$ is defined as:
\begin{equation}
\label{eq:overall_loss}
\begin{aligned}
L = \lambda_{1}L_{J3D} + \lambda_{2}L_{J2D} + \lambda_{3}L_{\theta} + \lambda_{4}L_{\beta} + \lambda_{5}L_{C}.
\end{aligned}
\end{equation}
In experiments, we set $\lambda_1 = \lambda_2 = 5.0$, $\lambda_3 = 1.0$, $\lambda_4 = 1\mathrm{e}{-3}$ and $\lambda_5 = 1.0$. These hyper-parameters are obtained from grid search.

\section{Experiments}
\label{sec:experiments}

In this section, we first introduce the experiment setup. 
Then we discuss the design of key features of the prototypical memory. 
After that, we compare our method with previous state-of-the-art methods. 
In the end, we provide further analysis on the design of {\methodname} and typical failure cases.

\subsection{Experimental Setting}
In our experiments, we employ several 3D human datasets: Human3.6M~\cite{ionescu2014human3}, MPI-INF-3DHP~\cite{mehta2017monocular}, 3DPW~\cite{vonMarcard2018}, UP-3D~\cite{lassner2017unite}. 
To increase the generalization ability of the models, we also adopt several in-the-wild 2D datasets, including LSP-Extended~\cite{Johnson11}, MPII~\cite{andriluka20142d} and COCO~\cite{lin2014microsoft}.
In our experiments, we use Human3.6M, MPI-INF-3DHP, MPII, LSP-Extended and COCO for training. Our models are evaluated on Human3.6M, MPI-INF-3DHP, 3DPW and UP-3D.
In the following of this subsection, we first discuss the evaluation metrics used in our work. Then we give a brief introduction to each dataset. In the end, we provide the implementation details.

\newpara{Evaluation Metrics.}
In this paper, we mainly use mean per-vertex position error (MPVPE)~\cite{xu2019denserac,rong2019delving} as the evaluation metric. 
We believe MPVPE is more suitable than joint-based errors such as MPJPE, since they only consider joint positions while neglecting the joint rotations and body shapes.
For completeness, we also adopt mean per-joint position error (MPJPE) and PA-MPJPE, which is MPJPE after applied Procrustes Analysis~\cite{gower1975generalized}. 
These two metrics are used to compare with the previous works and evaluate on datasets without ground-truth SMPL parameters, such as MPI-INF-3DHP~\cite{mehta2017monocular}.

%

\begin{figure*}[t]
	\begin{center}
		\includegraphics[width=0.95\linewidth]{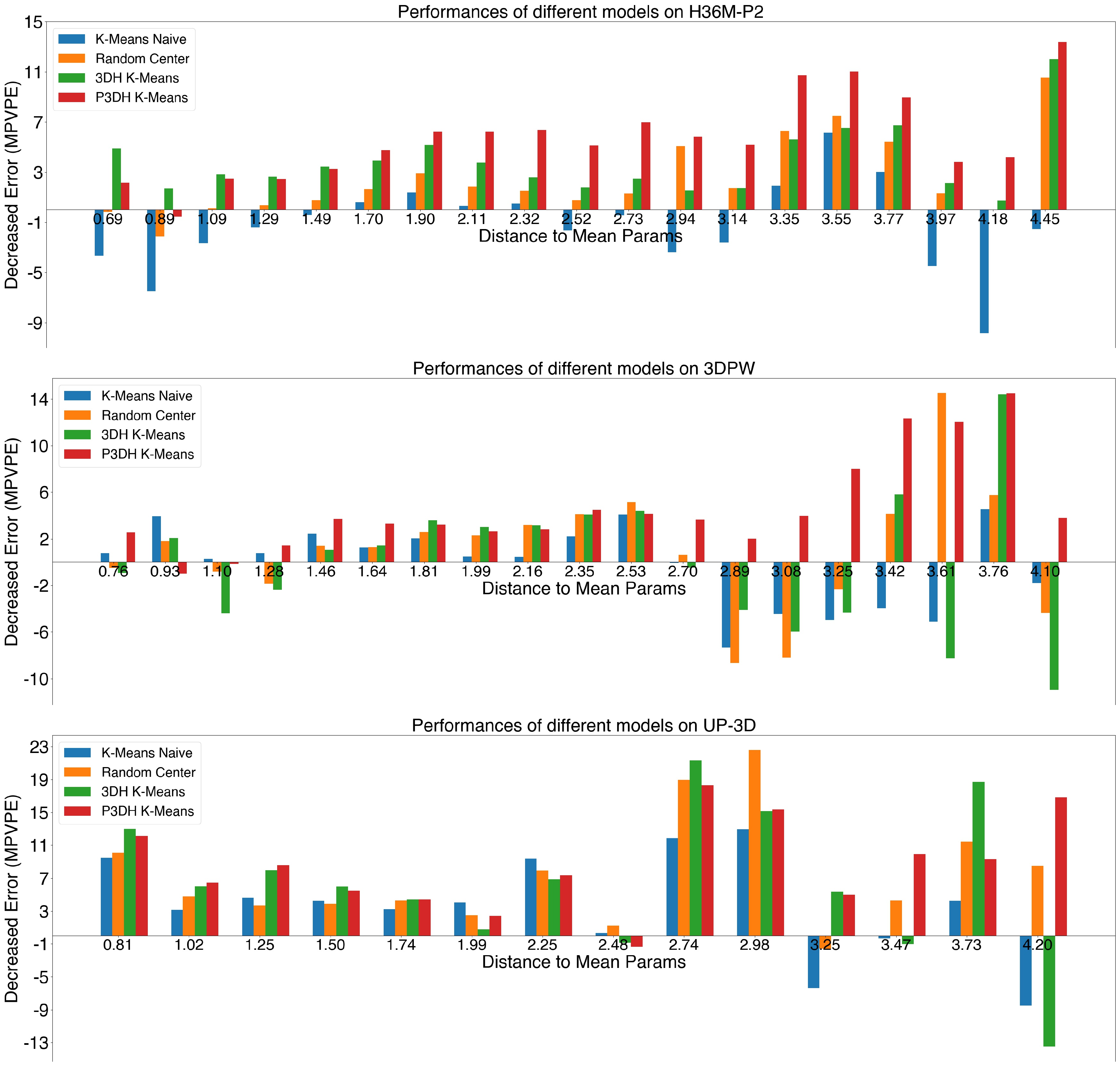}
	\end{center}
	\caption{\small \textbf{Ablation study}. We show the performances of four different variants of the proposed ~\methodname, namely K-Means Naive, Random Center, 3DH K-Means, P3DH K-Means. We evaluate on the evaluation set of Human3.6M~\cite{ionescu2014human3}, 3DPW~\cite{vonMarcard2018} and UP-3D~\cite{lassner2017unite}. Each evaluation set is divided into non-overlapping testing sets according to the samples' distances to the singular human prototype visualized in Fig.~\ref{fig:teaser} (a). The x-axis of each sub-figure is the samples' distances to the singular prototype. The y-axis is the models' decreased prediction errors compared to the baseline model (SPIN~\cite{kolotouros2019spin}).}
	\label{fig:ablation}
\end{figure*}

\begin{table}[t] \centering  \small %
	\caption{\textbf{Ablation Study.} Evaluating variants of prototypical memory. The evaluation metric is MPVPE.}
	\begin{tabular}{c|c|c|c}
		\hline
		Dataset $\rightarrow$                  		& \multirow{2}{*}{Human3.6M~\cite{ionescu2014human3}}    & \multirow{2}{*}{3DPW~\cite{vonMarcard2018}} & \multirow{2}{*}{UP-3D~\cite{lassner2017unite}} \\
		Methods $\downarrow$                   	    & & & \\    
		\hline
		SPIN~\cite{kolotouros2019spin}				& 78.7  & 117.8     & 119.3 \\
		\hline
		K-Means Naive						   	    & 79.1  & 116.6  & 115.1 \\
		Random Select								& 76.9  & 115.9  & 114.1 \\
		3DH K-Means								    & 75.4  & 115.8  & 113.5 \\
		P3DH K-Means							    & \textbf{73.6}  & \textbf{114.6}  & \textbf{112.6} \\   
		\hline
	\end{tabular}	
	\label{tab:ablation}
\end{table}

\begin{figure*}[t]
	\begin{center}
		\includegraphics[width=1.0\linewidth]{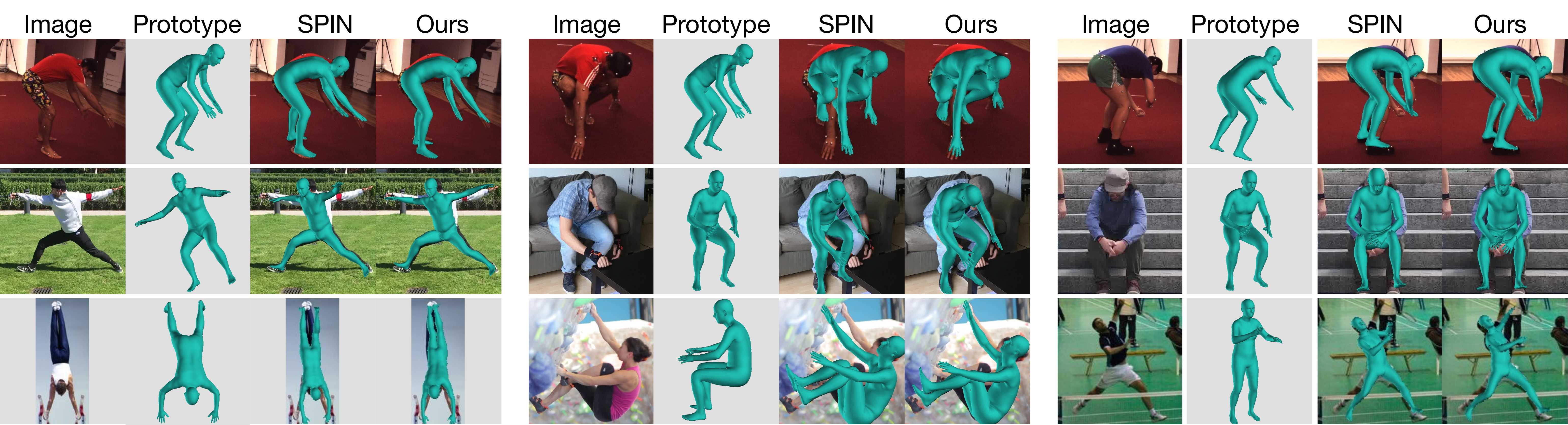}
	\end{center}
	\caption{\textbf{Qualitative comparison between {\methodname} and baseline model.} We qualitatively compare the {\methodname} with previous state-of-the-art methods, SPIN~\cite{kolotouros2019spin}. The images are selected from Human3.6M~\cite{ionescu2014human3} (Row 1), 3DPW~\cite{vonMarcard2018} (Row 2) and UP-3D~\cite{lassner2017unite} (Row 3).}
	\label{fig:qualitative_compare}
\end{figure*}

\begin{table*}[t]
	\small
	\begin{center}
		\caption{\textbf{Comparison with state-of-the-art methods (whole evaluation sets).} We compare proposed {~\methodname} with previous state-of-the-art methods on several widely used dataset including Human3.6M~\cite{ionescu2014human3}, 3DPW~\cite{vonMarcard2018}, UP-3D~\cite{lassner2017unite} and MPI-INF-3DHP~\cite{mehta2017monocular}. We use MPVPE, MPJPE and PA-MPJPE as the evaluation metrics.
			We mark the methods that use UP-3D in training with $*$. It is worth noting that our models are not trained on UP-3D.}
		\resizebox{2.0\columnwidth}{!}{
			\begin{tabular}{l|ccc|ccc|c|cc}
				\hline
				Dataset \& Metrics $\rightarrow$ & \multicolumn{3}{c|}{Human3.6M~\cite{ionescu2014human3}} & \multicolumn{3}{c|}{3DPW~\cite{vonMarcard2018}}    & UP-3D~\cite{lassner2017unite}  &\multicolumn{2}{c}{MPI-INF-3DHP~\cite{mehta2017monocular}}\\
				
				Methods $\downarrow$ 			 & MPVPE & MPJPE & PA-MPJPE         & MPVPE & MPJPE & PA-MPJPE    & MPVPE & MPJPE & PA-MPJPE\\
				\hline 
				Pavlakos~\etal~\cite{pavlakos2018learning}$^{*}$ & - & - & 75.9 & - & - & - & 117.7 & - & - \\
				CMR$^{*}$~\cite{kolotouros2019cmr} &	87.1 & 71.9 & 50.1 &	144.2 & 127.6 & 74.6 &	96.2 &	- & - \\
				DecoMR$^{*}$~\cite{zeng20203d} & 78.6 & 59.6 & \textbf{39.1} &	141.4 & 126.6 & 73.7 & 	\textbf{93.4} &	- & - \\		
				\hline
				HMR~\cite{kanazawa2018end} & - & - & 56.8 &	- & - & 81.3 &	- &	124.2 & 89.8 \\
				DenseRac~\cite{xu2019denserac} &- & - & 48.0 &	- & - & - & 	- &	114.2 & 83.5 \\
				SPIN~\cite{kolotouros2019spin} & 78.8 & 62.5 & 41.1 & 117.8 & 98.3 & 60.2 &	119.3 &	105.2 & 67.5 \\
				HKMR~\cite{georgakis2020hierarchical} & - & 60.0 & - & 	- & - & - &	- &	- & - \\		
				I2L-MeshNet~\cite{moon2020i2l} &- & \textbf{55.7} & 41.7 &	- & 95.4 & 58.6 & 	- &	- & -\\	
				Pose2Mesh~\cite{choi2020pose2mesh} &- & 67.9 & 49.9 &	- & \textbf{91.4} & 60.1 &	- &	- & - \\	
				\hline
				Ours & \textbf{73.6}  & 59.6 & 40.6 &	\textbf{114.6} & 94.6 & \textbf{58.4} &	\textbf{112.6} &	\textbf{98.2} & \textbf{62.8} \\
				\hline
			\end{tabular}
		}
		\label{tab:compare_sota_whole}
	\end{center}
\end{table*}

\begin{table*}[t]
	\begin{center}
		\caption{\textbf{Comparison with state-of-the-art methods (tail classes).} We compare with several state-of-the-art methods on tail classes on Human3.6M~\cite{ionescu2014human3}, 3DPW~\cite{vonMarcard2018} and UP-3D~\cite{lassner2017unite}. ``Tail-$x\%$'' means to only evaluate the data samples with the largest $x\%$ distances to the singular prototype as depicted in Fig.~\ref{fig:teaser} (a). The evaluation metric is MPVPE.}
		\resizebox{1.5\columnwidth}{!}{
			\begin{tabular}{l|cc|cc|cc}
				\hline
				Dataset \& Metrics $\rightarrow$ & \multicolumn{2}{c|}{Human3.6M~\cite{ionescu2014human3}} & \multicolumn{2}{c|}{3DPW~\cite{vonMarcard2018}}    & \multicolumn{2}{c}{UP-3D~\cite{lassner2017unite}}  \\
				Methods $\downarrow$ 			  & Tail-5$\%$ & Tail-10$\%$          & Tail-5$\%$ & Tail-10$\%$ & Tail-10$\%$ & Tail-20$\%$ \\
				\hline 
				CMR$^{*}$~\cite{kolotouros2019cmr}  & 92.5 & 97.3  & 209.9 & 194.3 & 170.4 & 149.2 \\
				DecoMR$^{*}$~\cite{zeng20203d} & 84.9 & 88.2  & 184.3 & 175.1 & \textbf{162.0} & \textbf{142.5} \\
				\hline
				SPIN~\cite{kolotouros2019spin}  &  89.0 & 88.8  & 130.0 & 130.6 & 184.2 & 167.9 \\
				Ours						   & \textbf{80.0} & \textbf{81.1} & \textbf{124.9} & \textbf{126.4} & \textbf{172.1} & \textbf{158.6} \\
				\hline
			\end{tabular}
		}
		\label{tab:compare_sota_tail}
	\end{center}
\end{table*}

\begin{figure}[t]
	\begin{center}
		\includegraphics[width=\linewidth]{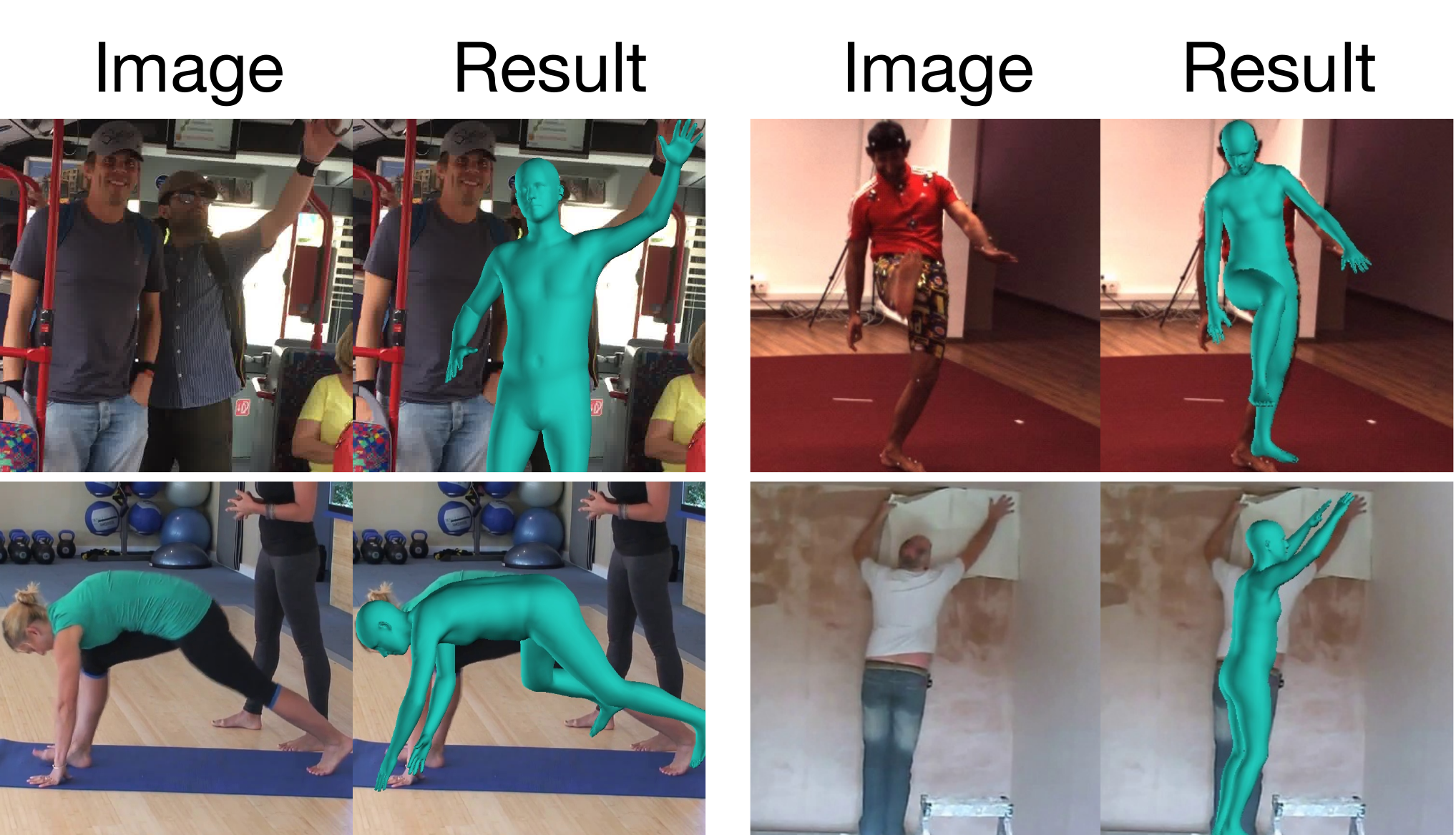}
	\end{center}
	\caption{\textbf{Failure cases.} Typical failure cases include occlusion, depth ambiguities, multiple person interaction and complicated backgrounds.}
	\label{fig:failure}
\end{figure}

\begin{figure*}[t]
	\begin{center}
		\includegraphics[width=\linewidth]{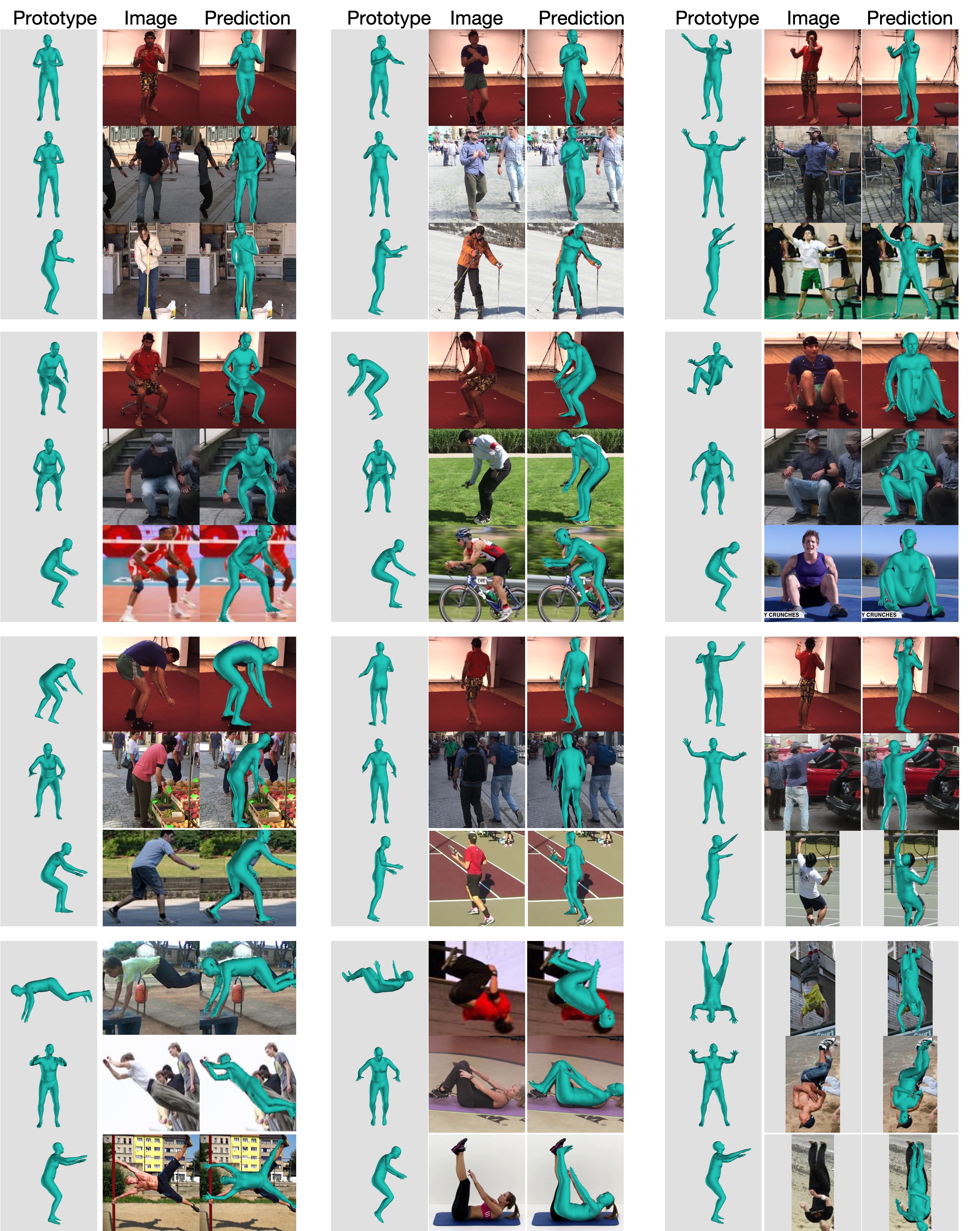}
	\end{center}
	\caption{\textbf{Qualitative Results.} Qualitative results of {\methodname} and P3DH K-Means. The images are selected from Human3.6M~\cite{ionescu2014human3}, 3DPW~\cite{vonMarcard2018} and UP-3D~\cite{lassner2017unite}. For each selected sample, we show the selected prototype, input images and {\methodname}'s prediction. Each prototype is rendered from original view, frontal view and side view.}
	\label{fig:qualitative}
\end{figure*}

\newpara{Human3.6M.}
Human3.6M~\cite{ionescu2014human3} is an indoor dataset with 3D joint annotations. We use Mosh~\cite{loper2014mosh} to collect SMPL parameters from 3D Mocap markers.
We train our models on subject S1, S5, S6, S7, and S8 and evaluate them on S9 and S11. 
Following the practice of previous works~\cite{kanazawa2018end,kolotouros2019spin}, we only evaluate on samples of frontal camera (camera 3). 
This evaluation strategy is often called ``Protocol-2'' in the literature.

\newpara{MPI-INF-3DHP.}
The dataset~\cite{mehta2017monocular} is captured with a multi-view camera system under a controlled environment. 
We use the subject S1 to S8 for training and evaluate our models on the evaluation set. 
The original dataset only provides 3D joints. 
In training, we also use the pseudo ground truth SMPL parameters provided by SPIN~\cite{kolotouros2019spin} through multi-view fitting. 

\newpara{3DPW.}
Data of 3DPW~\cite{vonMarcard2018} is captured under in-the-wild scenarios. The ground truth SMPL pose parameters are obtained from IMUs while shape parameters are obtained from the 3D scan. 
To enable fair comparisons with previous methods, we only use 3DPW's test set for evaluation and do not train our models on it.

\newpara{UP-3D.}
UP-3D~\cite{lassner2017unite} is composed of selected images from four 2D pose datasets. 
An extended SMPLify~\cite{bogo2016keep} is used to predict SMPL parameter on those images and then human annotators are asked to select the good samples. 
We only use UP-3D's test set for evaluation and do not train our model on it.

\newpara{2D Pose Datasets.}
To increase the generalization ability of the trained models, we additionally train our models on several in-the-wild datasets, including LSP-Extended~\cite{Johnson11}, MPII~\cite{andriluka20142d} and COCO~\cite{lin2014microsoft}.
These datasets only provide 2D landmark annotations. We use the SMPL fitting results provided by SPIN~\cite{kolotouros2019spin}.

\newpara{Implementation Details.}
All images are cropped around humans and scaled to $224 \times 224 $. 
The image encoder is composed of ResNet-50~\cite{he2016deep} with two additional fully connected layers, of which the output dimensions are both $1024$.
The classifier and parameter estimator are both fully connected layers, whose output dimensions are $K$ and $157$, respectively.
$K$ equals the number of 3D human prototypes in the prototypical memory, which is fixed to be $50$ throughout our experiments unless specified.
The $157$ dimensions are composed of shape parameters ($10$), pose parameters ($6\times 24$) and camera parameters ($3$). 
%
%
The batch size is $256$. Adam optimizer~\cite{kingma2015adam} with a learning rate $1\mathrm{e}{-4}$ is employed to train the models.

\subsection{Design of Prototypical Memory}

We use SPIN~\cite{kanazawa2018end}, which only adopts a single prototype, to serve as the baseline. 
Apart from reporting results on the overall evaluation set, we also split the evaluation set of Human3.6M~\cite{ionescu2014human3}, 3DPW~\cite{vonMarcard2018} and UP-3D~\cite{lassner2017unite} into un-overlapping subsets, according to samples' distances to the singular prototype, as depicted in Fig.~\ref{fig:teaser} (a).
We then report models' reconstruction errors (MPVPE) on each subset.

In our experiments, we evaluate four types of prototypical memory, including K-Means Naive, Random Center, 3DH K-Means and P3DH K-Means.
``K-Means Naive'' means directly applying the original K-Means algorithm on the SMPL parameters (pose parameters and shape parameters).
``Random Center'' uses initial centers, which are randomly selected from the samples, as the final cluster centers. 
The samples are assigned to clusters using the vertex-based distance without applying part-aware weighting.
``3DH K-Means'' refers to the P3DH K-Means without applying part-aware weighting.
These four different models share the same pipeline depicted in Fig.~\ref{fig:pipeline}. The training process and other hyperparameters, such as the number of clusters, are also the same.
The only differences are how the prototypical memories are built and how samples are assigned to the corresponding prototypes.
The evaluation results are shown in Fig.~\ref{fig:ablation}. 
The x-axis of each sub-figure is the samples' distances to the singular prototype depicted in Fig.~\ref{fig:teaser} (a). 
The y-axis is the models' error reduction compared to the baseline model (SPIN~\cite{kolotouros2019spin}). The error is MPVPE.
Besides, we also list evaluation results on the whole evaluation set Table~\ref{tab:ablation}.

It is revealed that P3DH K-Means achieves the best performance on nearly all the subsets of all the three datasets, especially on tail classes that are far from the singular human prototype.
3DH K-Means performs better than Random Center, proving that performing clustering to get better sample assignment is beneficial.
Furthermore, it is worth noting that directly performing naive K-Means clustering on SMPL parameters is not effective. It is even not comparable to Random Center. 
The experimental results prove the significance of performing the 3D human-specific clustering.

We further show qualitative results in Fig.~\ref{fig:qualitative_compare} to compare with the baseline model (SPIN~\cite{kolotouros2019spin}). 
The results show that our model can generate more precise results on challenging samples with rare poses such as squatting on the ground or uncommon views such as head upside down.
Results listed in Fig.~\ref{fig:qualitative_compare} are generated by {\methodname} using P3DH K-Means to build the prototypical memory.

\subsection{Main Results}
\label{sec:state_of_the_arts}

\newpara{Comparison with state-of-the-art methods.}
In this subsection, we compare our model with previous state-of-the-art methods. 
For methods with code released, we use their original code and models to obtain the evaluation results that are not included in their original papers.
For other methods, we report the original results listed in their papers.
We report evaluation results on both the whole evaluation sets and tail classes only. 
The results are listed in Table~\ref{tab:compare_sota_whole} for the overall evaluation results and Table~\ref{tab:compare_sota_tail} for tail class evaluation, separately. 
In Table~\ref{tab:compare_sota_tail}, ``Tail-$x\%$'' means to only evaluate samples with the largest $x\%$ distances to the singular prototype as depicted in Fig.~\ref{fig:teaser} (a).
The results in Table~\ref{tab:compare_sota_whole} show that our model achieves comparable performances with previous state-of-the-art methods on the whole evaluation sets. 
Furthermore, results in Table~\ref{tab:compare_sota_tail} show that our model achieves better results on tail classes.
These experimental results show that the proposed {\methodname} effectively improves the models' performances on challenging tail classes while achieving comparable or better performance on the samples.

\newpara{Qualitative Results.}
We show several qualitative results in Fig.~\ref{fig:qualitative}. 
It is demonstrated that the obtained prototypical memories contain both common poses and rare poses. 
Furthermore, the proposed {\methodname} can effectively select the matched prototypes for the input images and use them as the foundation to generate precise predictions.

\newpara{Failure Cases.} 
We show several typical failure cases in Fig~\ref{fig:failure}. 
As the images demonstrate, the model tends to fail when there are severe occlusions, ordinal depth ambiguities, multiple person interactions and challenging backgrounds.

\subsection{Further Analysis}

\begin{table}[t] \centering  \normalsize
	\caption{\small \textbf{Influence of the number of prototypes}. Experiment results of models using different number of prototypes.  We evaluation on the Human3.6M~\cite{ionescu2014human3}. The evaluation metric is MVPVE.}
	\setlength\tabcolsep{2pt}
	\begin{tabular}{c|c|c|c}
		\hline
		Dataset $\rightarrow$          			& H36M     & H36M   & H36M   \\
		Number of Prototypes $K$ $\downarrow$	&  All     &  Tail-$5\%$ & Tail-$10\%$  \\
		\hline
		5 & 75.9& 85.7& 86.9 \\
		10 & 74.9& 81.1& 81.7 \\
		50 & \textbf{73.6}& \textbf{80.0} & \textbf{81.1} \\
		100 & 76.4& 80.9 & 83.1 \\
		500 & 74.1& 86.5& 86.5 \\
		1000 & 74.8& 86.0& 86.3 \\
		\hline
	\end{tabular}	
	\label{tab:number_of_center}
\end{table}

\begin{table}[t] \centering   \normalsize
	\caption{\small \textbf{Influence of the weights of limb}. Experiment results of models using different weights for limb in performing part-aware weighting.  We evaluation on the Human3.6M~\cite{ionescu2014human3}. The evaluation metric is MVPVE.}
	\setlength\tabcolsep{2pt}
	\begin{tabular}{c|c|c|c}
		\hline
		Dataset $\rightarrow$          			& H36M     & H36M   & H36M   \\
		Weights of Limb $\downarrow$	&  All     &  Tail-$5\%$ & Tail-$10\%$  \\
		\hline
		1.0 & 75.5& 84.1& 83.4 \\
		3.0 & 76.6& 85.7& 86.9 \\
		5.0 & \textbf{73.6} & \textbf{80.0} & \textbf{81.1} \\
		7.0 & 73.6& 80.7& 82.0 \\
		10.0 & 75.5& 88.1& 88.4 \\
		\hline
	\end{tabular}	
	\label{tab:weight_of_limb}
\end{table}

In this subsection, we evaluate two designs of {\methodname} that influence the models' performance. 
1) The number of prototypes in prototypical memory. 
2) The weight of limbs in performing part-aware weighting.
The experiments are conducted on the evaluation set of Human3.6M~\cite{ionescu2014human3} with the same evaluation protocol defined in section~\ref{sec:state_of_the_arts}. 
The evaluation metric is MPVPE.
All the models reported in this subsection adopt P3DH K-Means to build the prototypical memory.
%

\newpara{Influence of the Number of Prototypes.}
We first study the influence of the number of prototypes (denoted as $K$). 
In our experiments, $K$ ranges from $5$ to $1000$. The results are listed in Table~\ref{tab:number_of_center}. 
The highest performance is achieved when $K$ equals to $50$. 
When $K$ is smaller than the optimal value, the capacity of prototype memory is not sufficient to provide the best-suited prototype for each data sample. 
When $K$ is larger than the optimal values, the data samples assigned to each prototype will decrease and becomes insufficient to train the conditional regressor.

\newpara{Influence of the Limb Weights.}
We next study the influence of the limb weights in performing part-aware clustering. 
%
%
We also do experiments to examine the influence of weights of other body parts.
It turns out the models' performances only differ slightly. Therefore, we only focus on the influence of limb weights in this subsection. 
The weights of other body part is set to be value shown in Fig.~\ref{fig:body_part}.
In our experiments, the weight of limb ranges from $1.0$ to $10.0$. The results are listed in Table~\ref{tab:weight_of_limb}. 
The highest performance is achieved when the weight is set to be $5.0$.
When the weight of limbs is smaller than the optimal value, the vertices of limbs are not assigned enough importance.
When the weight of limbs is larger than the optimal values, too much attention is paid to the limb, therefore affecting the learning of other parts.

\section{Conclusion}
In this paper, we propose Prototypical Memory Network (\methodname) to mitigate current single-image 3D human reconstruction models' suboptimal performance on rare poses.
The key of the proposed method is a prototypical memory that learns and stores a set of 3D human prototypes that can capture compact local distributions to represent both common and rare poses. 
It lifts the burden from the parameter regressor in coping with diverse poses.
While we mainly focus on the task of 3D human reconstruction in this paper, the notion of Prototypical Memory can be extended to other visual tasks that needs to handle data with multiple modes.



%
%

\ifCLASSOPTIONcaptionsoff
  \newpage
\fi



\bibliographystyle{IEEEtran}
\bibliography{IEEEabrv,long,LT3D}

\begin{thebibliography}{10}
\providecommand{\url}[1]{#1}
\csname url@samestyle\endcsname
\providecommand{\newblock}{\relax}
\providecommand{\bibinfo}[2]{#2}
\providecommand{\BIBentrySTDinterwordspacing}{\spaceskip=0pt\relax}
\providecommand{\BIBentryALTinterwordstretchfactor}{4}
\providecommand{\BIBentryALTinterwordspacing}{\spaceskip=\fontdimen2\font plus
\BIBentryALTinterwordstretchfactor\fontdimen3\font minus
  \fontdimen4\font\relax}
\providecommand{\BIBforeignlanguage}[2]{{%
\expandafter\ifx\csname l@#1\endcsname\relax
\typeout{** WARNING: IEEEtran.bst: No hyphenation pattern has been}%
\typeout{** loaded for the language `#1'. Using the pattern for}%
\typeout{** the default language instead.}%
\else
\language=\csname l@#1\endcsname
\fi
#2}}
\providecommand{\BIBdecl}{\relax}
\BIBdecl

\bibitem{kolotouros2019spin}
N.~Kolotouros, G.~Pavlakos, M.~J. Black, and K.~Daniilidis, ``Learning to
  reconstruct {3D} human pose and shape via model-fitting in the loop,'' in
  \emph{IEEE International Conference on Computer Vision}, 2019.

\bibitem{ionescu2014human3}
C.~Ionescu, D.~Papava, V.~Olaru, and C.~Sminchisescu, ``{Human3.6M}: Large
  scale datasets and predictive methods for {3D} human sensing in natural
  environments,'' \emph{IEEE Transactions on Pattern Analysis and Machine
  Intelligence}, vol.~36, no.~7, pp. 1325--1339, 2014.

\bibitem{vonMarcard2018}
T.~von Marcard, R.~Henschel, M.~Black, B.~Rosenhahn, and G.~Pons-Moll,
  ``Recovering accurate {3D} human pose in the wild using {IMUs} and a moving
  camera,'' in \emph{European Conference on Computer Vision}, 2018.

\bibitem{zeng20203d}
W.~Zeng, W.~Ouyang, P.~Luo, W.~Liu, and X.~Wang, ``{3D} human mesh regression
  with dense correspondence,'' in \emph{IEEE Conference on Computer Vision and
  Pattern Recognition}, 2020, pp. 7054--7063.

\bibitem{kanazawa2018end}
A.~Kanazawa, M.~J. Black, D.~W. Jacobs, and J.~Malik, ``End-to-end recovery of
  human shape and pose,'' in \emph{IEEE Conference on Computer Vision and
  Pattern Recognition}, 2018.

\bibitem{guler2019holopose}
R.~A. Guler and I.~Kokkinos, ``Holopose: Holistic {3D} human reconstruction
  in-the-wild,'' in \emph{IEEE Conference on Computer Vision and Pattern
  Recognition}, 2019, pp. 10\,884--10\,894.

\bibitem{kolotouros2019cmr}
N.~Kolotouros, G.~Pavlakos, and K.~Daniilidis, ``Convolutional mesh regression
  for single-image human shape reconstruction,'' in \emph{IEEE Conference on
  Computer Vision and Pattern Recognition}, 2019.

\bibitem{loper2015smpl}
M.~Loper, N.~Mahmood, J.~Romero, G.~Pons-Moll, and M.~J. Black, ``{SMPL}: A
  skinned multi-person linear model,'' \emph{Transcation on Graphics}, 2015.

\bibitem{rogez2019lcr}
G.~Rogez, P.~Weinzaepfel, and C.~Schmid, ``Lcr-net++: Multi-person {2D} and
  {3D} pose detection in natural images,'' \emph{tpami}, vol.~42, no.~5, pp.
  1146--1161, 2019.

\bibitem{mehta2017monocular}
D.~Mehta, H.~Rhodin, D.~Casas, P.~Fua, O.~Sotnychenko, W.~Xu, and C.~Theobalt,
  ``Monocular {3D} human pose estimation in the wild using improved {CNN}
  supervision,'' in \emph{International Conference on 3D Vision}.\hskip 1em
  plus 0.5em minus 0.4em\relax IEEE, 2017, pp. 506--516.

\bibitem{lassner2017unite}
C.~Lassner, J.~Romero, M.~Kiefel, F.~Bogo, M.~J. Black, and P.~V. Gehler,
  ``Unite the people: Closing the loop between {3D} and {2D} human
  representations,'' in \emph{IEEE Conference on Computer Vision and Pattern
  Recognition}, 2017.

\bibitem{pavlakos2018learning}
G.~Pavlakos, L.~Zhu, X.~Zhou, and K.~Daniilidis, ``Learning to estimate {3D}
  human pose and shape from a single color image,'' in \emph{IEEE Conference on
  Computer Vision and Pattern Recognition}, 2018.

\bibitem{xu2019denserac}
Y.~Xu, S.-C. Zhu, and T.~Tung, ``Denserac: Joint {3D} pose and shape estimation
  by dense render-and-compare,'' in \emph{IEEE International Conference on
  Computer Vision}, 2019.

\bibitem{rong2019delving}
Y.~Rong, Z.~Liu, C.~Li, K.~Cao, and C.~C. Loy, ``Delving deep into hybrid
  annotations for {3D} human recovery in the wild,'' in \emph{IEEE
  International Conference on Computer Vision}, 2019.

\bibitem{moon2020i2l}
G.~Moon and K.~M. Lee, ``I2l-meshnet: Image-to-lixel prediction network for
  accurate {3D} human pose and mesh estimation from a single {RGB} image,'' in
  \emph{European Conference on Computer Vision}, 2020.

\bibitem{georgakis2020hierarchical}
G.~Georgakis, R.~Li, S.~Karanam, T.~Chen, J.~Kosecka, and Z.~Wu, ``Hierarchical
  kinematic human mesh recovery,'' in \emph{European Conference on Computer
  Vision}, 2020.

\bibitem{zanfir2020weakly}
A.~Zanfir, E.~G. Bazavan, H.~Xu, B.~Freeman, R.~Sukthankar, and
  C.~Sminchisescu, ``Weakly supervised {3D} human pose and shape reconstruction
  with normalizing flows,'' in \emph{European Conference on Computer Vision},
  2020.

\bibitem{goodfellow2014generative}
I.~Goodfellow, J.~Pouget-Abadie, M.~Mirza, B.~Xu, D.~Warde-Farley, S.~Ozair,
  A.~Courville, and Y.~Bengio, ``Generative adversarial nets,'' in
  \emph{Advances in Neural Information Processing Systems}, 2014, pp.
  2672--2680.

\bibitem{densepose}
R.~Alp~Güler, N.~Neverova, and I.~Kokkinos, ``Densepose: Dense human pose
  estimation in the wild,'' in \emph{IEEE Conference on Computer Vision and
  Pattern Recognition}, 2018.

\bibitem{choi2020pose2mesh}
H.~Choi, G.~Moon, and K.~M. Lee, ``Pose2mesh: Graph convolutional network for
  {3D} human pose and mesh recovery from a {2D} human pose,'' in \emph{European
  Conference on Computer Vision}, 2020.

\bibitem{kipf2017semi}
T.~N. Kipf and M.~Welling, ``Semi-supervised classification with graph
  convolutional networks,'' in \emph{International Conference on Learning
  Representation}, 2017.

\bibitem{bogo2016keep}
F.~Bogo, A.~Kanazawa, C.~Lassner, P.~Gehler, J.~Romero, and M.~J. Black, ``Keep
  it {SMPL}: Automatic estimation of {3D} human pose and shape from a single
  image,'' in \emph{European Conference on Computer Vision}, 2016.

\bibitem{zhou2019continuity}
Y.~Zhou, C.~Barnes, J.~Lu, J.~Yang, and H.~Li, ``On the continuity of rotation
  representations in neural networks,'' in \emph{IEEE Conference on Computer
  Vision and Pattern Recognition}, 2019, pp. 5745--5753.

\bibitem{markley2007averaging}
F.~L. Markley, Y.~Cheng, J.~L. Crassidis, and Y.~Oshman, ``Averaging
  quaternions,'' \emph{Journal of Guidance, Control, and Dynamics}, vol.~30,
  no.~4, pp. 1193--1197, 2007.

\bibitem{Johnson11}
S.~Johnson and M.~Everingham, ``Learning effective human pose estimation from
  inaccurate annotation,'' in \emph{IEEE Conference on Computer Vision and
  Pattern Recognition}, 2011.

\bibitem{andriluka20142d}
M.~Andriluka, L.~Pishchulin, P.~Gehler, and B.~Schiele, ``{2D} human pose
  estimation: New benchmark and state of the art analysis,'' in \emph{IEEE
  Conference on Computer Vision and Pattern Recognition}, 2014.

\bibitem{lin2014microsoft}
T.-Y. Lin, M.~Maire, S.~Belongie, J.~Hays, P.~Perona, D.~Ramanan,
  P.~Doll{\'a}r, and C.~L. Zitnick, ``Microsoft {COCO}: Common objects in
  context,'' in \emph{European Conference on Computer Vision}, 2014.

\bibitem{gower1975generalized}
J.~C. Gower, ``Generalized procrustes analysis,'' \emph{Psychometrika},
  vol.~40, no.~1, pp. 33--51, 1975.

\bibitem{loper2014mosh}
M.~Loper, N.~Mahmood, and M.~J. Black, ``Mosh: Motion and shape capture from
  sparse markers,'' \emph{Transcation on Graphics}, vol.~33, no.~6, p. 220,
  2014.

\bibitem{he2016deep}
K.~He, X.~Zhang, S.~Ren, and J.~Sun, ``Deep residual learning for image
  recognition,'' in \emph{IEEE Conference on Computer Vision and Pattern
  Recognition}, 2016.

\bibitem{kingma2015adam}
D.~P. Kingma and J.~Ba, ``Adam: A method for stochastic optimization,''
  \emph{International Conference on Learning Representation}, 2015.

\end{thebibliography}
\end{document}